# Key Features of the Coupled Hand-operated Balanced Manipulator (HOBM) and Lightweight Robot (LWR)

Yang Zhang[1,2], Vigen Arakelian[1,2], Baptiste Véron[3] and Damien Chablat[1]

[1] LS2N UMR 6004, 1 rue de la Noë, BP 92101,
F-44321 Nantes, France
[2] Mecaproce / INSA-Rennes
20 av. des Buttes de Coesmes, CS 70839,
F-35708 Rennes, France
[3] IRT Jules Verne, Chemin de Chaffault
F-44340 Bouguenais, France

`vigen.arakelyan@insa-rennes.fr`

**Abstract.** The paper deals with coupled systems including hand-operated balanced manipulators and lightweight robots. The aim of such a cooperation is to displace heavy payloads with less powerful robots. In other term, in the coupled system for handling of heavy payloads by a HOBM an operator is replaced by a LWR. The advantages of the coupled HOBM and LWR are disclosed and the optimal design of the cooperative workspace is discussed. Behavior of the coupled system in a static mode when the velocities the HOBM are limited does not present any special problems. In this mode, the inertial forces are significantly lower than the gravitational one. The payload is completely balanced by the HOBM and the LWR assumes the prescribed displacements with low load. However, in a dynamic mode, the HOBM with massive links creates additional loads on the LWR, which can be significant. The present study considers a method for determination of inertia effects of the HOBM on the LWR. The given numerical simulations show the significant increasing of the input torques due to the inertia forces of the HOBM. Behavior of the HOBM with cable lift and the LWR is also examined.

## 1 Introduction

Workers in industries such as manufacturing and assembly, frequently manipulate heavy objects. However, manual processing is often repetitive and becomes tedious, it reduces efficiency and leads to back pains, injuries and musculoskeletal disorders. It is obvious that traditional robot installations can offer several benefits compared to manual operation: improved repeatability, increased precision and speed. However, industrial robots still have many weaknesses compared to humans. For example, currently industrial robots have a limited ability to perceive their surroundings, which requires costly safety arrangements in order to avoid serious injury. These safety arrangements are particularly important and costly when working with installations of large and powerful industrial robots. It is obvious that serial robots have a poor payload-to-weight ratio. For a six-degrees-of-freedom general-type serial robot, it is less than 0.15 [0]. For example, a robotic arm handling an object of $50 kg$ must have a weight of at least $350\ kg$. The purchase, installation and operation of such a robot is quite expensive. In addition, the heaviness of the robot and of the payload complicates the dynamics of the system, making it difficult to move accurately and quickly. This becomes especially noticeable during assembly processes, when heavy parts must be installed on a surface with guiding pins. In such a case, the robotic arm has to move smoothly and any sudden movement may damage the mechanical surface of the part. Such a task is not easy to achieve. Thus, autonomous manipulation does not always provide expected reliability and flexibility.

## 2     Advantages of the coupled HOBM and LWR

In our view, balancer – robot systems such as power assist robotic systems may be perfectly used for heavy object manipulation. The combination of motion programming of a lightweight robot and simplicity of a hand-operated balanced manipulator (HOBM) may make the system far better than the application of an individual robot arm.

Let's now consider HOBM applications. The HOBM is a handling system with a simple mechanical actuator in which the manipulated object in any position of the workspace is balanced. Such a state of constant balance allows displacements of heavy objects to be achieved manually. The advantages of these manipulators relatively to industrial robots are the simplicity of their construction and their low cost. They have a great weight-carrying capacity and a very large workspace. The implantation of HOBM in existing production line is very simple without the need of important additional surfaces, special auxiliary devices or essential reorganization of the production lay-out. Different approaches and solutions devoted to the design and balancing of HOBM have been developed and documented [2]-[13].

These manipulators have found a broad application in several fields of industry where it is necessary to carry out mechanization of heavy manual work. The production of HOBM can be found in several countries, for example, «Dalmec» (Italy), «Balaman» (Japan), «Conco-Balancer» (USA, Fig. 1), «Auto-Balancer» (Germany), «Triom» (Czech Republic), «Yaplex» (UK, Fig. 2), «CEM» (France), etc. The use of such manipulators includes a mode of operation when mechanical devices and humans cooperate to hold and move objects.

Let's now consider a HOBM and a LWR cooperation for handling of heavy parts. In other words, let us consider a new coupled system in which an operator is replaced by a LWR (Fig. 3).

It should be noted that such an approach has drawn attention of researchers and engineers since a long time. The new method of coordinative control of a robot arm and HOBM [14] has been developed in order to automate batch production processes, including heavy parts assembly. Fig. 4 shows the coupled system called the «Hitachi Progress Robot» having a payload capacity of $300N$.

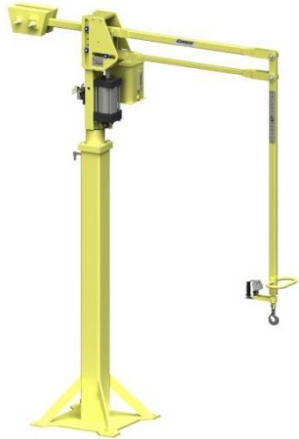

**Fig. 1.** Conco-Balancer (USA).

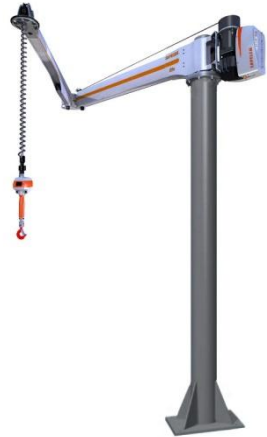

**Fig. 2.** Sapalem Manipulator (UK).

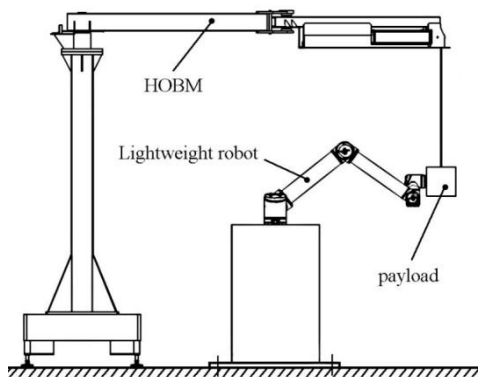

**Fig. 3.** LWR and HOBM cooperation for handling of heavy parts.

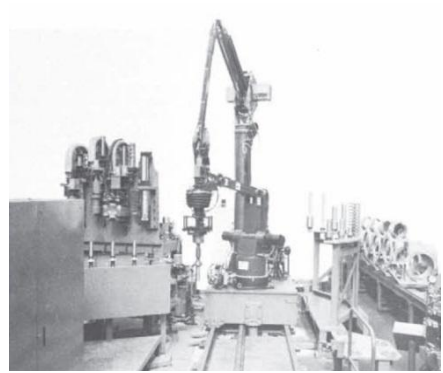

**Fig. 4.** Hitachi Progress Robot (Japon).

The characteristics of the robot arm and the HOBM used in this coupled system have been following: a five degrees of freedom robot arm with payload capacity: $100N$, maximum velocity: $1m/s$, repeatability: $\pm 0.2mm$ and a four degrees of freedom HOBM with payload capacity: $1500N$ and maximum velocity: $0.2m/s$. The performance of the Hitachi Progress Robot was a payload capacity of $300N$, maximum vertical velocity: $0.2m/s$, maximum horizontal velocity: $0.4m/s$, vertical repeatability: $\pm 0.3mm$ and horizontal repeatability: $\pm 0.5mm$.

However, nowadays, taking into account the great capabilities of manual lightweight robots, which allow human intervention to control and guide the payload, such a cooperation becomes much more efficient since it does not exclude the possibility of having a human in the workspace of a robot. Thus, it can reduce costs for space and safety measures as shared space is possible.

## 3     Optimal design of the cooperative workspace

When designing coupled systems, it is necessary to keep in mind that they consist of two units with different characteristics. However, some of their parameters can be modified during the cooperation of these units. One of the first is to consider structural compliance, i.e. any movement of the payload carried out by means of a LWR must be accompanied by a HOBM. If there is a discrepancy between the movements of these two units, the coupled system will be blocked.

The design of units can also be modified. The vertical axis of the HOBM is usually able to rotate over 360 degrees to provide the largest workspace (Fig. 5).

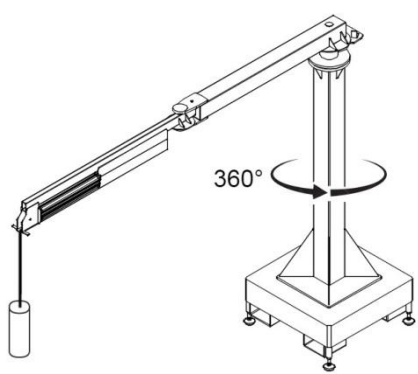 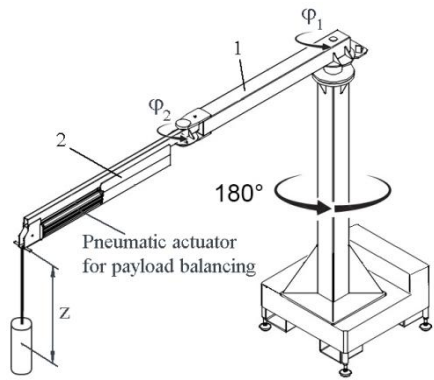

**Fig. 5.** Workspace of the HOBM.          **Fig. 6.** Modified HOBM for cooperation with LWR.

However, in the coupled system with joint workspace, the HOBM uses only a small portion of its reachable space due to the workspace of the LWR, which are usually limited in its volume. Thus, the design of the HOBM can be modified to be adapted to the new conditions. The vertical axis of the HOBM can be brought closer to the LWR, and the static balance can be saved with a counterweight on the opposite side of the coupled system workspace (Fig. 6). Such an arrangement is more optimal in terms of the cooperative workspace. Practical implementation of such modifications may look as shown in Fig.7.

It should also be noted that the cooperative workspace of the coupled system must be collision and singularity free. Let's illustrate with an example. In Fig. 8 is presented a coupled system in which the HOBM is in the singular configuration. In the case of the conventional usage of the HOBM, this configuration does not present any inconvenience, since the operator will not move the payload in the radial direction. He will remove the load from the singular position of the manipulator and then perform the necessary movements. However, in a coupled system, it is indispensable to take this into account when planning a trajectory of the LWR, since such a movement cannot be performed. Thus, it is necessary to avoid not only the singular configurations of the LWR but also of the HOBM.

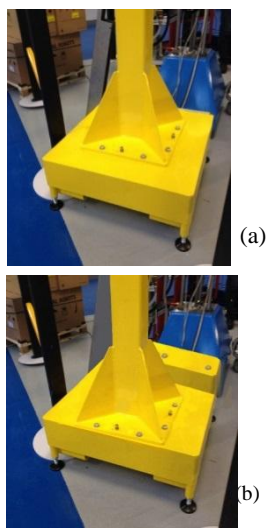

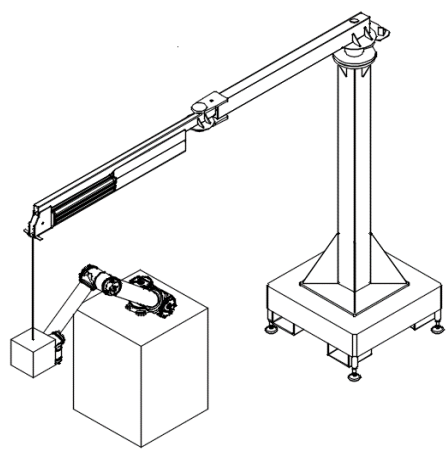

**Fig. 7.** Frame of the usual HOBM (a) and its modified version (b) for balancer - lightweight robot cooperation.

**Fig. 8.** Coupled system in which the HOBM is in the singular configuration.

## 4   Behavior of the coupled system in dynamic mode

Behavior of the coupled system in a static mode when the velocities of links the HOBM are limited does not present any special problems. In this mode, the inertial forces are significantly less than the gravitational. The payload is completely balanced by the HOBM and the LWR assumes low loads. However, in case of increasing of accelerations, the inertia forces also increase and, respectively, the efficiency of gravitational balancing decreases. Therefore, the HOBM with massive links creates additional loads on the LWR. Let's consider the loads of a HOBM on the lightweight manipulator.

In the coupled system, the payload is moved by the LWR. Thus, knowing the vector $\boldsymbol{\theta}$ of joint angles of the LWR, the vector $\dot{\mathbf{x}}$ of Cartesian velocities can be determined from:

$$\dot{\mathbf{x}} = \mathbf{J}_{l.m.}\dot{\boldsymbol{\theta}}$$

where, $\mathbf{J}_{l.m.}$ is the Jacobian matrix of the LWR.

The displacements and velocities of the payload are the input parameters for the HOBM. Hence,

$$\dot{\boldsymbol{\varphi}} = \mathbf{J}_{\text{HOBM}}^{-1}\dot{\mathbf{x}}$$

where, $\mathbf{J}_{\text{HOBM}}$ is the Jacobian matrix of the HOBM and $\boldsymbol{\varphi}$ is the vector of joint angles of the HOBM.

Thus, knowing the vector $\boldsymbol{\varphi}$ of joint angles, the inertial forces and moments of the HOBM can be determined and the loads $\boldsymbol{\tau}_{\text{HOBM}}$ due to these forces and moments in the joints will also be established.

Now, the force-torque vector $\mathbf{F}_{\text{HOBM}}$ acting on the payload due to the HOBM can be expressed by the relationship:

$$\mathbf{F}_{\text{HOBM}} = \mathbf{J}_{\text{HOBM}}^{-T}\boldsymbol{\tau}_{\text{HOBM}}$$

This force-torque vector will create supplementary torques in the joints of the lightweight manipulator. The dynamic equations describing the input torques of the lightweight manipulator with the HOBM action can be expressed as:

$$\boldsymbol{\tau} = \boldsymbol{\tau}_{l.m.} + \mathbf{J}_{l.m.}\mathbf{F}_{\text{HOBM}}$$

where, $\boldsymbol{\tau}_{l.m.} = \mathbf{M}(\boldsymbol{\theta})\ddot{\boldsymbol{\theta}} + \mathbf{V}(\boldsymbol{\theta},\dot{\boldsymbol{\theta}})\dot{\boldsymbol{\theta}} + \mathbf{G}(\boldsymbol{\theta})$.

Let's consider an illustrative example through a coupled system including the HOBM having the structure similar to the manipulator given in Fig.8, i.e. with two rotating links (1), (2) and a rigid telescopic axis (3) for vertical displacements of the payload and a six-degrees-of-freedom general-type LWR. The mass-inertia characteristics of units are given in tables 1 and 3, where for the $i$-th link the mass is denoted as $m_i$, the length denoted as $l_i$, the location of the center mass $S_i$ from the previous joint denoted as $r_{S_i}$, the axial moment of inertia relative to the center of mass of the link denoted as $I_{S_i}$, the transverse moment of inertia relative to the center of mass of the link denoted as $I_{x_i}$, $I_{y_i}$ and $I_{z_i}$. The Denavit–Hartenberg parameters of the LWR are given in table 2. For numerical simulations the following input motion has been applied:

$$\theta_1(t) = \begin{cases} \theta_1^i + \frac{1}{2}t^2\alpha & (0 \leq t \leq \tau) \\ \theta_1^i + (t - \frac{\tau}{2})\omega & (\tau \leq t \leq t_f - \tau) \\ \theta_1^f - \frac{1}{2}(t_f - t)^2\alpha & (t_f - \tau \leq t \leq t_f) \end{cases}$$

with $\theta_1^i = -40°$, $\theta_1^f = 40°$, $\tau = 0.2s$, $t_f = 2s$, $\omega = \left(\theta_1^f - \theta_1^i\right)/\left(t_f - \tau\right)$, $\alpha = \omega/\tau$, $\theta_2 = -45°$, $\theta_3 = 90°$, $\theta_4 = -225°$, $\theta_5 = 90°$ and $\theta_6 = 0°$.

**Table 1.** Mass-inertia characteristics of the HOBM

| Link $i$ | $m_i$ (kg) | $l_i$ (m) | $r_{S_i}$ (m) | $I_{S_i}$ (kgm²) |
|---|---|---|---|---|
| 1 | 30.97 | 1.4 | 0.57 | 9.28 |
| 2 | 23.56 | 1.5 | 0.74 | 5.21 |
| 3 | 2.13 | 0.6 | 0.3 | 0.06 |

**Table 2.** Denavit-Hartenberg parameters of the LWR

| Joint $i$ | $\theta_i$ (rad) | $a_i$ (m) | $d_i$ (m) | $\alpha_i$ (rad) |
|---|---|---|---|---|
| 1 | $\theta_1$ | 0 | 0.1273 | $\pi/2$ |
| 2 | $\theta_2$ | -0.612 | 0 | 0 |
| 3 | $\theta_3$ | -0.572 | 0 | 0 |
| 4 | $\theta_4$ | 0 | 0.163941 | $\pi/2$ |
| 5 | $\theta_5$ | 0 | 0.1157 | $-\pi/2$ |
| 6 | $\theta_6$ | 0 | 0.0922 | 0 |

**Table 3.** Mass-inertia characteristics of the LWR

| Link $i$ | $m_i$ (kg) | $l_i$ (m) | $r_{S_i}$ $[x_{S_i}, y_{S_i}, z_{S_i}]$ (m) | $I_{x_i}$ (kgm²) | $I_{y_i}$ (kgm²) | $I_{z_i}$ (kgm²) |
|---|---|---|---|---|---|---|
| 1 | 1.35 | 0.038 | [0, 0.0116, 0.0786] | $4.62 \times 10^{-3}$ | $5.40 \times 10^{-3}$ | $4.88 \times 10^{-3}$ |
| 2 | 3.82 | 0.612 | [0, 0.251, 0.0844] | $1.20 \times 10^{-1}$ | $8.08 \times 10^{-1}$ | $6.96 \times 10^{-1}$ |
| 3 | 2.04 | 0.572 | [0, 0.258, 0.0566] | $8.03 \times 10^{-3}$ | $2.96 \times 10^{-1}$ | $2.90 \times 10^{-1}$ |
| 4 | 0.32 | 0.164 | [0, 0.009, 0.0463] | $5.35 \times 10^{-4}$ | $4.79 \times 10^{-4}$ | $4.07 \times 10^{-4}$ |
| 5 | 0.32 | 0.116 | [0, 0.010, 0.0464] | $5.37 \times 10^{-4}$ | $4.82 \times 10^{-4}$ | $4.06 \times 10^{-4}$ |
| 6 | 0.07 | 0.092 | [0, 0, 0.0126] | $5.72 \times 10^{-5}$ | $5.95 \times 10^{-5}$ | $6.57 \times 10^{-5}$ |

The input torques of the LWR are given in Fig. 9: (a) without HOBM (b) taking into account inertial forces of the HOBM. One can see from Fig. 9, the inertia forces of the HOBM significantly increase the loads on the actuators of the LWR. This example clearly shows that the influence of the inertia of the HOBM can be significant during high-speed movements of the LWR. Thus, it is necessary to take into account when calculating loads on the actuators, not only the inertia of the payload but also the inertia of the HOBM.

Behavior of the coupled system in a dynamic mode is completely different for a HOBM with a cable lift assuming vertical movements of the payload (Fig. 8). In this case, the dynamic loads on the LWR occur in the form of oscillations of the HOBM at the end of the operating cycle when the LWR stops. These oscillations essentially depend on the friction in the joints of the HOBM. Such a dependence is shown in Figure 10. One of the ways of reduction of these unwanted oscillations is the increasing the friction on the joints of the HOBM.

Unfortunately, increasing the friction in the joints of the HOBM creates a drag force that the robot has to overcome for any movement. Then, two efforts are in competition: 1) the inertial effort due to the oscillations of the HOBM, and 2) the drag effort created by the friction. Reducing 1) means increasing 2). Thus, an optimal friction has to be found such that the drag effort is minimized and the oscillation reduction remains efficient.

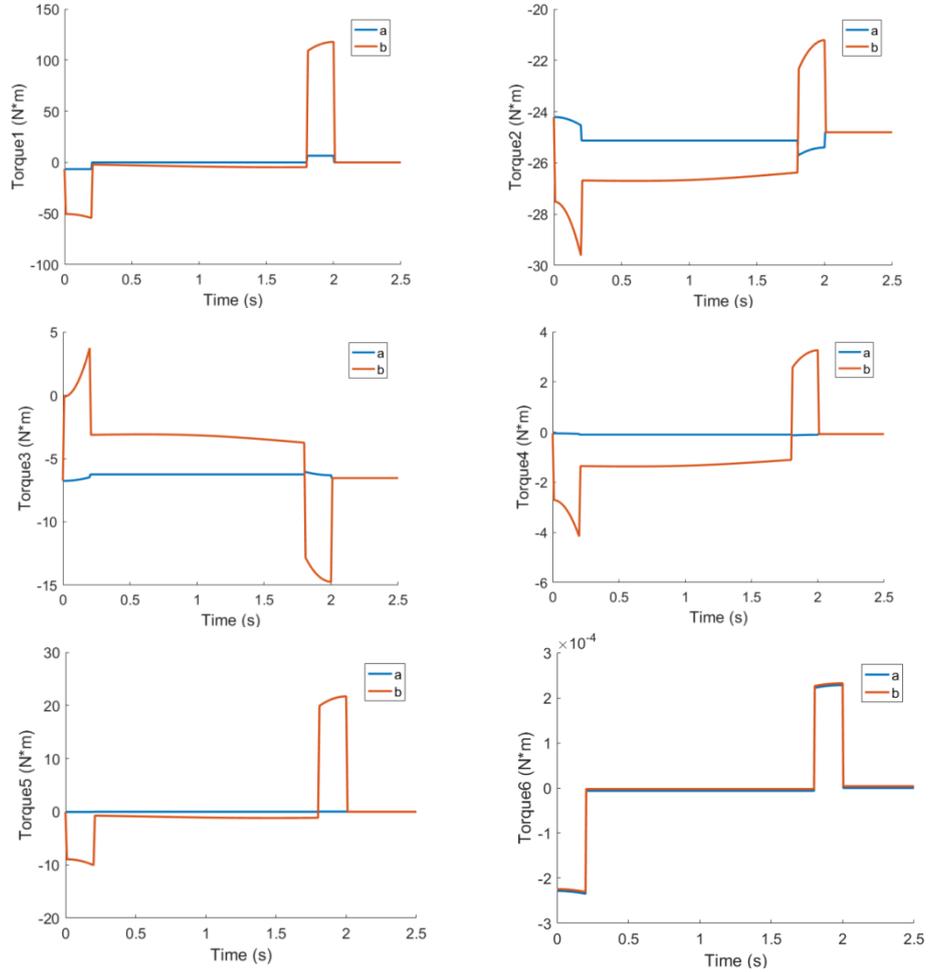

**Fig. 9.** Input torques of the LWR: (a) without HOBM (b) taking into account inertial forces of the HOBM.

Such an optimization highly depends on the application for the coupled system. Indeed, the efforts will mainly depend on the friction, of course, but they will also depend on the accelerations and trajectories imposed by the lightweight robot and the payload.

One possible method to perform this optimization is to run a design of experiment (DOE). In our case, the measure studied is the maximal effort applied on the robot. The parameters of the DOE are 1) the friction in the joints of the HOBM, 2) the mass of the payload, 3) the acceleration. A response surface design is chosen (central-composite) in order to build a meta-model of our system. This meta-model is a quadratic model that gives us the influence of each parameter on the measured effort.

Then, this model can be used to tune the parameters. For instance, Fig. 11 shows how to tune the parameters to obtain a maximal effort on the robot not exceeding 120 N.

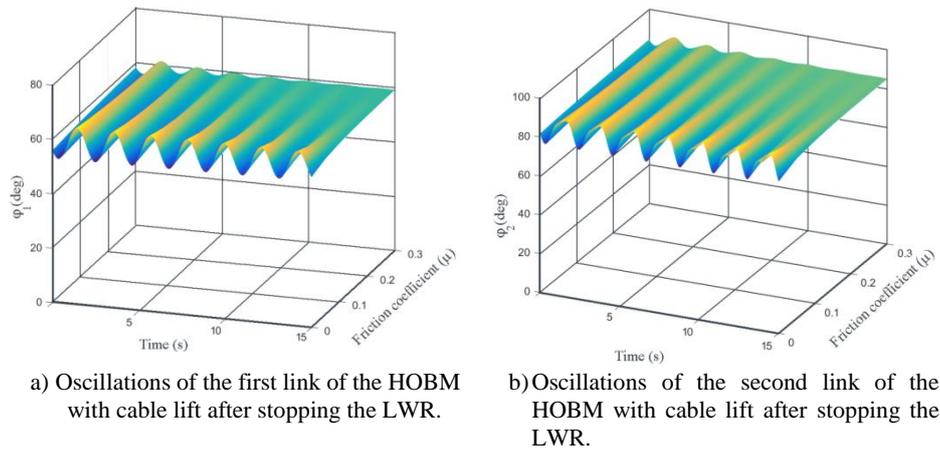

a) Oscillations of the first link of the HOBM with cable lift after stopping the LWR.

b) Oscillations of the second link of the HOBM with cable lift after stopping the LWR.

**Fig. 10.** Oscillations of the rotating links of the HOBM with cable lift after stopping the LWR.

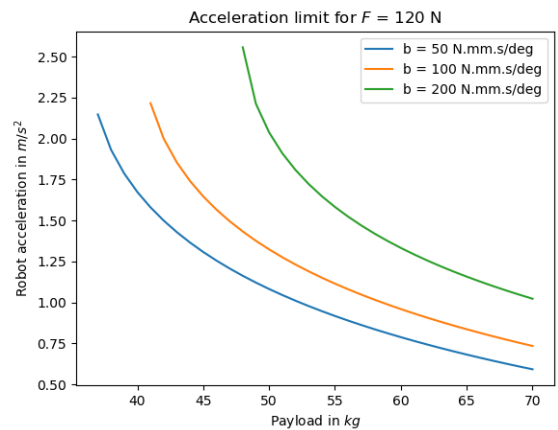

**Fig.11.** Acceleration limit for F = 120 N.

## 5   Conclusions

Manual handling of loads involves the use of the human body to lift and carry loads. Most manufacturing systems require some manual handling tasks. When performed incorrectly or excessively, these tasks may expose workers to fatigue and injury. A variety of techniques and tools exist for automatic handling of heavy loads. Light-weight robots and hand-operated balanced manipulators cooperation is an effective way. In this case, for handling of heavy payloads by a HOBM an operator is replaced by a LWR. Taking into account the great capabilities of manual lightweight robots, which allow human intervention to control and guide the payload, such a cooperation is efficient since it does not exclude the possibility of having a human in the workspace of a robot.

The present study deals with the key features of the coupled HOBM and LWR. There are shown the design particularities of the cooperative workspace, the need to consider singular configurations of the HOBM, the method for determination of inertia effects of the HOBM on the LWR. It is revealed that in a dynamic mode the HOBM with massive links creates additional loads on the LWR, which can be significant. The given numerical simulations show the important increasing of the input torques due to the inertia forces of the HOBM. Behavior of the HOBM with cable lift and the LWR is also examined. It is disclosed that there are significant oscillations of the rotating links of the HOBM with cable lift after stopping the LWR. It is proposed to eliminate these unwanted oscillations by controlling friction in the joints of the HOBM.